# The Forecastability of Underlying Building Electricity Demand from Time Series Data

Mohamad Khalil
School of Engineering
Newcastle University
Newcastle upon Tyne, UK
m.khalil2@newcastle.ac.uk

A. Stephen McGough
School of Computing
Newcastle University
Newcastle upon Tyne, UK
stephen.mcgough@ncl.ac.uk

Hussain Kazmi
Dept. of Electrical Engineering
KU Leuven
Leuven, Belgium
hussainsyed.kazmi@kuleuven.be

Sara Walker
School of Engineering
Newcastle University
Newcastle upon Tyne, UK
sara.walker@ncl.ac.uk

*Abstract*— Forecasting building energy consumption has become a promising solution in Building Energy Management Systems for energy saving and optimization. Furthermore, it can play an important role in the efficient management of the operation of a smart grid. Different data-driven approaches to forecast the future energy demand of buildings at different scale, and over various time horizons, can be found in the scientific literature, including extensive Machine Learning and Deep Learning approaches. However, the identification of the most accurate forecaster model which can be utilized to predict the energy demand of such a building is still challenging.

In this paper, the design and implementation of a data-driven approach to predict how forecastable the future energy demand of a building is, without first utilizing a data-driven forecasting model, is presented. The investigation utilizes a historical electricity consumption time series data set with a half-hour interval that has been collected from a group of residential buildings located in the City of London, United Kingdom.

The proposed methodology mainly consists of four steps: firstly, we utilized four data-driven approaches (daily and weekly naive, Light Gradient Boosting Machine, and Linear Regression) to predict the day-ahead of building energy demand, and generate target labels of interest. The four forecasting approaches have been evaluated by using the Root Mean Squared Error, and Mean Absolute Error. Secondly, two feature extraction packages have been utilized to convert each of the building electricity demand time series into a feature matrix. Thirdly, we added the label of interest (i.e. best forecaster model) to each element of the extracted feature matrix. Finally, we utilized a classification data-driven approach (i.e. Random Forest) on the extracted feature datasets to predict how forecastable the future energy demand of such a building is.

The experimental results demonstrate that it is possible to make a prior estimates about the forecastability of certain electricity demand time series of such a building.

## I. Introduction

Building Energy Management Systems (BEMSs) are equipped with smart meters that record the energy consumption of buildings. This historical data can be utilized by data-driven approaches, for example, Machine Learning (ML) and Deep Learning (DL), to predict the future energy demand of the buildings. This could aid the planning of the wider energy system, and also enable improvement to the energy management of such a building. Furthermore, according to Khalil et al. [1], the engineering task of forecasting building energy consumption is indispensable when one takes into account that the building stock is one of the biggest consumers of energy worldwide. Therefore, the topic of building energy analysis and performance has attracted a significant attention in the energy sector. Recently, there has been large success of data-driven approaches in solving several challenging tasks in different domains. Researchers in the field of building energy analysis and performance are also starting explore to data-driven models for predicting the future energy demand of buildings. The majority of previous research works in the context of this study have focused on forecasting the future energy demand of buildings over different time horizons. However, predicting the time series property of the underlying building energy demand is still a challenging task. Therefore, the aim of this research is to propose a new data-driven framework to predict how forecastable the future energy demand of such a building is, without first training the data-driven forecasting model. This research work is organized as follows. Section II Introduces related research studies. Section III describes the data characteristics and studied buildings. Section IV presents the methodology and implementation process of the forecastability framework. Then Section V shows the assessment and evaluation of the proposed data-driven models. Finally, the conclusions and recommendations for future works are given. In the next section, we present related research works in the context of this study.

## II. Related Work

As energy demand reduction is a global imperative, researchers have extensively utilized several approaches including physics-based, and data-driven to predict the energy demand [2]. The reader is referred to [3] where more details about physics-based approaches for building energy analysis and performance can be found. The data-driven approaches, e.g. ML and DL, utilize the following input features to construct a forecaster model which are: (I) the historical time series energy demand, (II) meteorological information such as outdoor dry-bulb temperature, and wind speed, (III) calendar information, for example, hour of the day, day of the week, and (IV) operational data such as occupancy density [4].

Extensive ML and DL approaches have been used, , Wang et al. [5] utilised 11 input features that have been collected from two educational buildings located at the University of Florida, including ambient, occupancy and time related data, to predict the hourly energy consumption using a Random Forest (RF) model. The authors demonstrated the performance superiority of RF to predict building energy demand when comparing with other conventional ML approaches such as Support Vector Regression (SVR), and Regression Tree (RT). Chen and Tan [6] used a hybrid data-driven approach which is a SVR combined with Multi-resolution Wavelet Decomposition (MWD) to predict the hourly electric demand intensity in two buildings: a mall and a hotel. The results demonstrated the introduction of MWD to SVR can improve the prediction performance for the two case studies. Moreover, previous research studies have compared SVR with other ML approaches such as Artificial Neural Network (ANN) [7] to predict the seasonal electricity consumption in Turkey. The comparison results showed seasonal SVR approach was able to predict the electricity consumption better than the ANN, indicating its potential to be a suitable approach for predicting building energy consumption effectively.

In addition, state-of-the-art DL approaches have attracted great attention [8]. For example, Rahmana at al. [9] employed a DL approach which is a Recurrent Neural Network (RNN) with LSTM units. The proposed approach was utilized to predict the electricity consumption for multiple cases, e.g. commercial and residential buildings over medium-to-long term time horizons. The authors tested the effectiveness of the RNN by using the following input features: outdoor weather, schedule-related, and historical load profiles. The research results showed that the proposed approach is a promising solution for solving the forecasting building energy consumption task over medium to long-term time horizons owing to its ability to capture long-term temporal dependencies in time-series data.

More recently, Wang et al. [10] proposed a Long-Short-Term-Memory (LSTM) approach to predict the energy demand for a group of office buildings located in Alaska, United States. The authors compared the performance of the LSTM approach with three other prediction ML approaches, including SVR, RF, and Multi-layer Perceptron (MLP). The experimental results showed that LSTM performed better than comparative ML approaches with respect to Mean Absolute Percentage Error (MAPE).

This related work demonstrates that researchers have employed a plethora of data-driven forecasting approaches. However, there are extremely few research studies focused on the identification of the most suitable data-driven approach which can be used to forecast the energy demand of a particular building. Therefore, this research work proposes a data-driven forecastability framework to fill the knowledge gap between data science communities, and building professionals. Inspired by the work that was done by Hyndman et al. [11], this research study employs two features extortion packages which are: (I) Interpretable Feature Extraction of Electricity Loads (IFEEL), e.g. domain-informed features [12], and (II) Nixtla, e.g. domain-agnostic features [13], on building energy time series data. Our research is different from AutoML system. AutoML system identifies for a given dataset the best ML model to use.

Whilst, in this research, we are learning from latent features that extracted from the dataset as to which ML model to utilize. The main contribution of this research study is to present an end-to-end data-driven framework that can predict the time series property of building energy demand (i.e. the forecastability of the underlying time series). The studied buildings, and its data characteristics will be explained in more details in the next section.

### III. Experimental design

#### A. Tested Building

A group of 500 residential buildings located in the City of London, United Kingdom, were utilized as a case study to evaluate the effectiveness of the proposed forecastability framework, this dataset was collected by UK Power Network [14]. The residential buildings in the UK power network project were recruited as a balanced sample representative of the Greater London population. Electricity consumption readings from these household were taken at 30 mins intervals. Normally, the City of London features a temperate oceanic climate with cold and windy winter, while summers tend to be warmer and sunnier. The average winter temperature is $7^{\circ}$ C, while the average temperature in summer is $18^{\circ}$ C.

#### B. Data

The proposed data-driven forecasting models were trained and tested on the electricity consumption data that was obtained from the project. The dataset consists of 36 input features, and one output feature. The input features of this dataset include: (I) ambient features (i.e., outdoor dry-bulb temperature with an average of $8.6^{\circ}$ C in the winter and an average of $15.9^{\circ}$ during the summer, dew point, wind speed, relative humidity), (II) historical electricity demand, and (III) time related features due to their solid correlations with the occupants' profile (i.e., hour of the day, parts of the day, workday type, day of the week, month of the year, and season of the year). The meteorological information was collected by using the visual crossing API [15], which provides historical and forecast weather data. The output feature is the predicted hourly building electricity consumption. To ensure the integrity and consistency of the datasets [16], a data preprocessing method was conducted which replaced the missing data points using their nearest complete data points. Furthermore, some households had a mean total electricity consumption of zero, which were discarded from the analysis. For the sake of clarity, we present the full time series, total daily and mean electricity consumption for one randomly chosen building, i.e. MAC000010, from the case study in Fig. 1. While Fig. 2. and Fig. 3. show the hourly electricity consumption, heat map of the electricity usage of this chosen building. Based on the heat map, it can be seen that the hours between 18:00 and 22:00 exhibit higher electricity consumption. The following subsection demonstrates the data-driven forecastability framework in details.

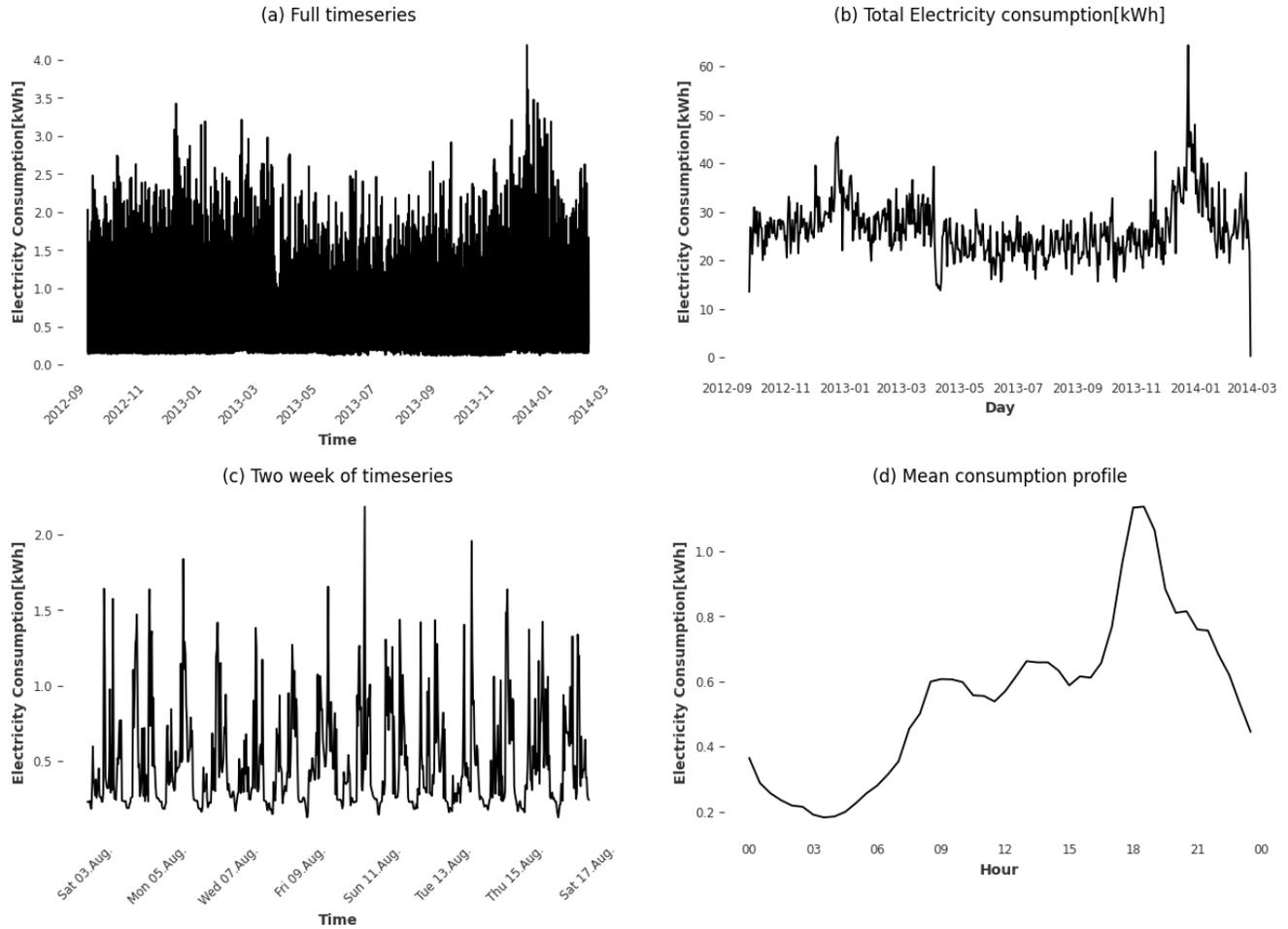

Fig. 1. (a) represents the full historical electricity consumption in half-hour interval, (b) shows the average daily electricity consumption, (c) shows two weeks of time series, and (d) represents the mean hourly consumption of the chosen building, i.e. MAC000010.

*C. Forecastability Framework*

The overall forecastability workflow in this study can be seen in Fig. 4., and it consists of the following steps:

(I) Forecast the electricity load of the studied buildings for the day-ahead by using several data-driven models, given historical electricity load data, and other covariates features.

(II) The domain informed, and domain agnostic features of the studied buildings are extracted through the use (IFEEL) package [12], and tsfeatures functionality from the Nixtla package [13] respectively. These feature extraction techniques are based on the idea of transforming a single time series with dimensions of (1, ntimestamps) into a vector of features with shape (1, nfeatures). Typical applications for extracting features from electricity demand time series data are similar buildings detection, and forecastability prediction. In this work, the methodology of creating the features matrix from the electricity demand time series data sets, e.g. ifeel, Nixtla, and a combination of both, is based on work by Canaydin et al. [27]. The authors created a feature matrix from multiple case studies including the UK low carbon dataset by using the following package: (I) ifeel: domain-informed feature, this package takes a single electricity demand time series to produces a set of 95 physical features such as mean, max, and skewness, and (II) Nixtla: domain-agnostic features, this package extracts a set of 42 features such as series length, stability and sparsity from a collection of time series. A comprehensive explanation about the list of domain-agnostic features, and domain-informed features can be found in [28] and [27] respectively.

(III) The annotation process that includes adding the label/target of interest, i.e. the best forecaster model which has the lowest Root Mean Square Error (RMSE) value, for each time series in the extracted feature matrix (i.e. ifeel, Nixtla, and a combination of both) in order to train and evaluate a data-driven classification model that will be utilized to learn the mapping between the extracted input features, and the target labels.

(IV) Employ a RF classification model for the time series property prediction task, i.e. identification of the most suitable driven approach.

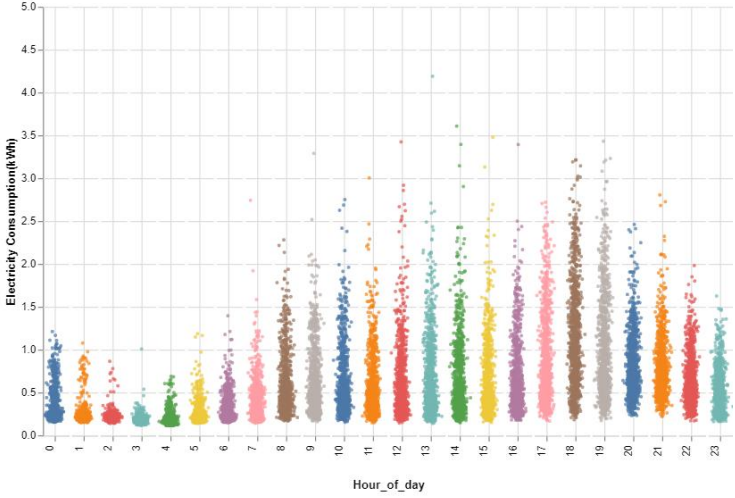

Fig. 2. The hourly electricity consumption profile [kWh].

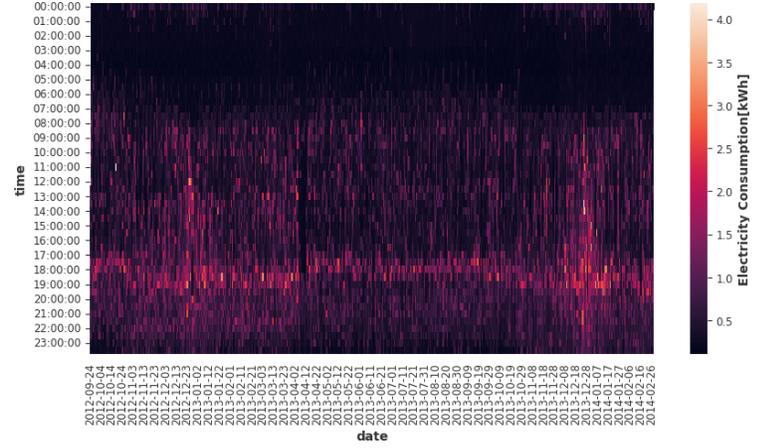

Fig. 3. Heat map of the electricity consumption [kWh].

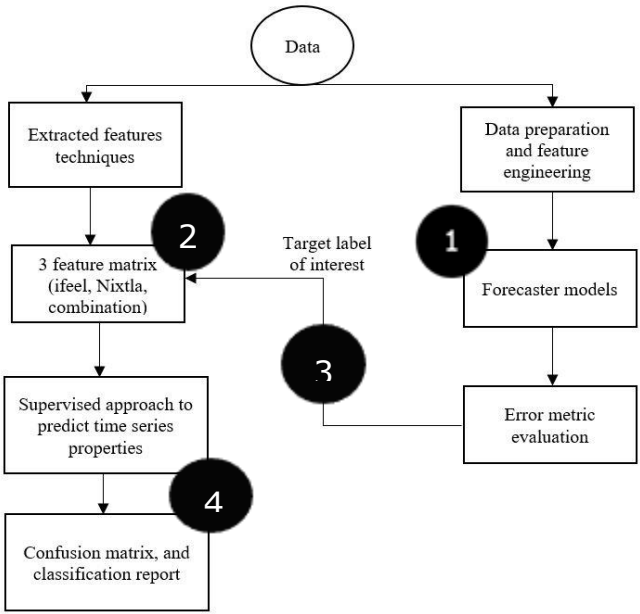

Fig. 4. The proposed data-driven forecastability framework.

## IV. FORECASTING METHODOLOGY

In this section, the process of forecasting and modelling building electricity consumption is introduced. It mainly focuses on how to predict the day-head electricity demand of the studied buildings by using data-driven approaches. The models involved in the research work, i.e., persistence models, Linear Regression (LR) and LightGBM, are introduced in subsections A, B, and C respectively.

The forecasting process of the proposed data-driven models in this study is as follows:

I. As described earlier, the required predictor variables to fit the data-driven models are 36 features including the covariates, i.e., meteorological, and time-related features, and the historical electricity demand of such a building.

II. With regard to lagged values of the covariates features, the lookback and look-forward window was equal to 48 time steps. While the lagged values of the historical electricity measurement were set to 336 timesteps, e.g. previous week, for the weekly naïve, MLR, and LightGBM models, and the 48 timesteps, e.g. the previous day, for the daily naïve model.

III. Predict the day-ahead electricity consumption of the studied buildings.

IV. Evaluate the prediction performance of the proposed data-driven models according to the following accuracy metrics which are (I) Root Mean Squared Error (RMSE) that measures the difference between all the predicted and actual values. RMSE values can range from zero to positive infinity, Low RMSE values indicate that the data-driven model has good predictions capability. RMSE metric is widely used in forecasting applications because it has the same units as the dependent variable. Mean Absolute Error (MAE) that calculates the average absolute residuals between the actual and precited values.

$$\text{MAE} = \frac{1}{n} \sum_{i=1}^{n} |y_i - \hat{y}_i| \quad (1)$$

$$\text{RMSE} = \sqrt{\frac{\sum_{i=1}^{n}(y_i - \hat{y}_i)^2}{n}} \quad (2)$$

Where $y_i$ is the actual electricity consumption value, $\hat{y}_i$ is the predicted value, and $n$ = the number of measurements for the comparison.

Keeping in mind that we implemented the same forecasting methodology on all the studied 500 buildings. All the proposed data-driven forecasting models have been implemented using the DARTS API [17]. The preliminary knowledge of the forecasting models is presented in the following paragraphs.

### A. Naïve/Persistence Forecast

The persistence model is one of the simplest data-driven approaches which is often used as a performance benchmark against more sophisticated ML and DL approaches [18]. These models, in general, have been utilized for solving several forecasting tasks in different domains such as finance [19] and energy [20], [21] due to their simplicity and computational efficiency. Regarding a forecasting building electricity consumption task, the persistence model predicts the next data point into the future at time $t+1$ by utilizing the last available data point at time $t$ as equation (3) shown.

$$y_{T+h} = y_t \quad (3)$$

Where the notation $y_{T+h}$ represents the future observation, while $y_t$ is the historical data. In this research work two persistence models were employed which are daily, and weekly naïve.

### B. Linear Regression Forecast

LR based methods belong to the Regression Analysis (RA) family [1]. A simple LR method is utilized to study the underlying correlation between one predictor feature $X$, e.g. meteorological information, and one dependant feature $Y$, e.g. the future electricity demand. This then creates the relation between the predictor, and dependant features. There are several types of RA based methods which can be used For example, Multiple Linear Regression (MLR) [22] is when the regression model describes the relationship between multiple predictor features and one dependant feature; Autoregressive Integrated Moving Average (ARIMA) [23], where the assumptions of ARIMA models with respect to time series data are stationary, i.e. the mean and variance value of time series are consistent over time. In this work, the MLR method has been selected to predict the day-ahead electricity demand of the studied buildings. The equation of the MLR model can be seen in formula (4).

$$\hat{y}_t = \beta_0 + \beta_1 X_1 + \beta_2 X_2 + \ldots + \beta_n X_n + \varepsilon \quad (4)$$

where $y$ is the dependant feature, $X$ is the predictor feature, $B$ is the intercept, and $\varepsilon$ is the error.

### C. LightGBM Forecast

LightGBM was first introduced in 2017 by Ke et al. [24], the main advantages of this model over its competitors such as Gradient Boosting Decision Tree (GBDT) is its computational and memory efficiency, and scalability when working with high-dimensional data. According to [24] gradient-based one-side sampling, and exclusive feature bundling are the novel ideas of LightGBM. One of the key difference between LightGBM and traditional gradient boosting models is that the decision trees in LightGBM are extended leaf-wise rather than checking all of the preceding leaves for each new leaf [25]. As reported by Zhang et al. [26], the training procedure of regressor LightGBM to generate the final prediction value can be described in the following steps: (I) initialize prediction for the first Decision Tree (DT) in the ensemble, (II) the latter DT learns how to fit the residuals from the former trees, (III) the final prediction of LightGBM is the summation of all the outcomes from the DTs in the ensemble.

## V. PREDICTING TIME SERIES PROPERTY

Although many data-driven models are available, the choice of the specific model to forecast the building energy consumption is not an easy task, and depends strongly on the data characteristics. Therefore, this research study fills the gap in the existing literature by using a supervised data-driven approach that learns the mapping function between the target labels of interest (i.e. best performing forecaster model), and three extracted feature matrix to determine the most suitable forecasting model for future unseen electricity demand time series data. The following subsection present an RF model that is used to predict the underlying electricity demand time series.

### Random Forest

RF is an ensemble data-driven model that operates by aggregating several DTs to process the final prediction [29]. The central theme of an RF model is that this model is trained through the random variable selection and bagging approach, i.e. Parallel Homogeneous, where each tree in the forest is trained on different sub-set of samples and features from the original training set. The final prediction outcome of an RF is constructed according to the task; in a regression task, the outcome is the mean prediction for all the trees in the forest, while in a classification task is the mode class which has been selected by the majority of the trees within the forest. There are several advantages of RFs over its competitors such as traditional tree based models including: (I) RFs are less prone to overfitting on the training set, (II) they are robust to high-dimensional input data, and (III) can reduce the overall variance of the model's prediction. The equation of an RF model for a classification task can be seen in formula (5), where $T_i$ is the total number of the trees, and $C$ is the target label.

$$RF(x) = max_c \sum_{i=1}^{N}(T_i(x) = C) \quad (5)$$

In this research study an RF classification model has been utilized to predict the best forecaster model for a given instance of extracted time series feature from all the constructed datasets, (i.e. ifeel, Nixtla, and combination of both). The full details about the experimental results of this study is introduced in next section.

## VI. RESULTS ANALYSIS

In this section, we present the results, beginning with assessing the forecasting capability of the data-driven models with respect to predicting the day-ahead electricity consumption of the studied 500 buildings, to evaluating the classification performance of an RF model to predict the best forecaster model.

*A. Forecasting Evaluation*

Figures 5 and 6 show the boxplot results for the RMSE and MAE values respectively for predicting the day-ahead electricity consumption of the studied buildings. These figures indicate that on average LR and LightGBM models are more capable than persistence models with respect to the forecasting accuracy. While these models are more efficient than benchmark models, they still have rather long training times (4-5 minutes compared to 1-2 minutes per household). With respect to creating the target label of interest, we choose the best forecaster model for each of the extracted time series features according to several error metrics such as RMSE, MAE, rMAE. Due to page length limit, only the RMSE results are reported, i.e. the lowest RMSE value indicates the model outperformed its competitors. The percentage of the target label of interest can be seen in Fig. 7. It is shown LR and LightGBM were the best forecaster model for (217) and (272) buildings respectively. On the other hand, both of the persistence models scored the lowest RMSE value for just 10 buildings. The possible reason for the relativity unsatisfied performance of the naïve models is that these models cannot consider external feature during the modelling process. These results confirm that there is no one-size-fits-all data-driven approach that can be utilized to solve different case studies. Therefore, the need to estimate the forecastability of a certain electricity demand time series is inevitable.

*B. Forecastability Evaluation*

In this section, an RF-based classification approach is proposed to predict the underlying time series building electricity demand. The proposed approach is applied to learn the mapping between input features ($X$), and the target label of interest ($Y$) to predict the best forecaster model for future unseen observations. As a result, we applied three separate RF classifiers on three different input feature matrix (ifeel, Nixtla, and combination of both). To construct the RF models, we used 75% of the extracted feature matrix as the training set, and the remaining 25% as a testing set. The grid search method has been utilized to find the optimal set of parameters for the proposed models, and the performances were assessed by using the cross validation approach [30]. The classification models were implemented by using the scikit-learn API [31].

In this study, two error matrix were used to evaluate and compare the performance of the proposed models which are: (I) confusion matrix, and (II) classification report [32]. A confusion matrix summarizes the classification results by displaying the numbers from predicted and actual values. The outputs of the confusion matrix, True Positive (TP), and True Negative (TN) are the numbers of positive class and negative class have been predicted accurately by the classification model. While False Positive (FP), and False Negative (FN) refer to the total numbers of positive and negative classes which have been predicted inaccurately. The results of the confusion matrices of the ifeel, Nixtla, combination models are shown in the Figures 8 and 10 respectively. Based on the classification reports of the experiments, it can infer that the RF model that has been applied on the combination dataset has the best classification accuracy with 74%, while domain-informed and domain-agnostic models led to accuracy score of 72% and 68% respectively. Therefore, those findings prove the advantage of combining the ifeel, Nixtla feature matrix together.

Furthermore, we utilized the RF feature importance technique on the three proposed scenarios, to identify the features that have the greatest contributions toward the final outcome of the classification models. The top (5) features, along with their contributions for the proposed RF models is shown in Figures. 11 and 13. In light of this analysis, we can notice, four out of the top five features of the combination model are domain-informed features. This is due to the fact that the domain-informed package, i.e. ifeel, has been neatly devised for electricity time series data, while the domain-agnostic package, i.e. Nixtla, is an off-the-shelf solution which is developed for any type of time series data.

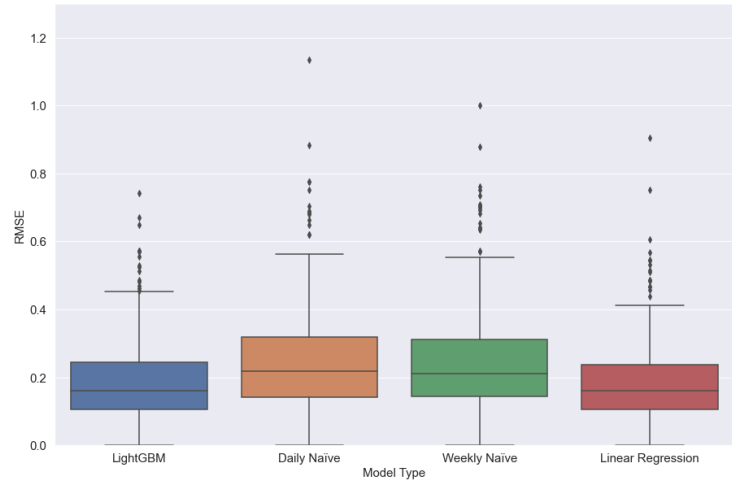

Fig. 5. RMSE values for the 500 studied buildings.

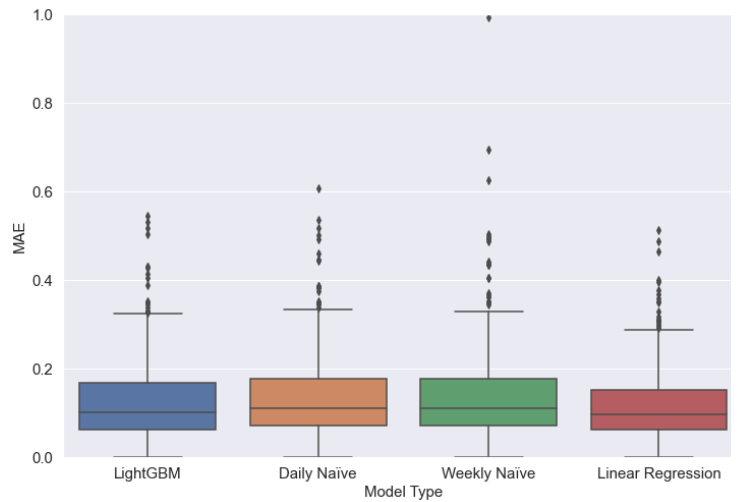

Fig. 6. MAE values for the 500 studied buildings.

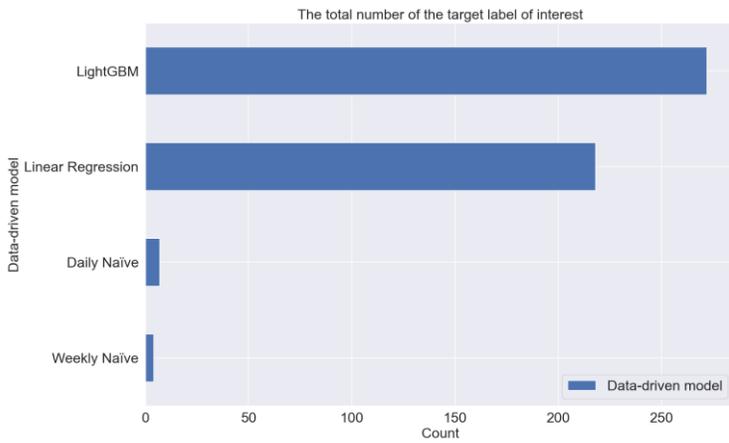

Fig. 7. The total number of the target label of interest.

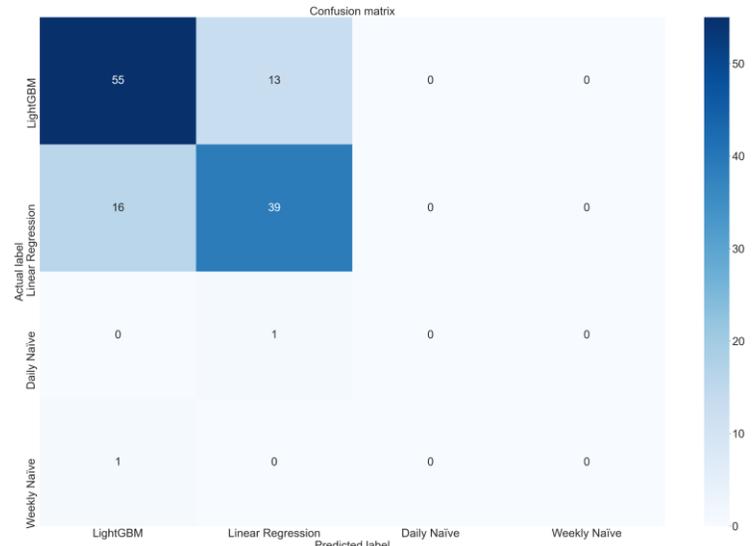

Fig. 10. Confusion matrix of the combination classification model.

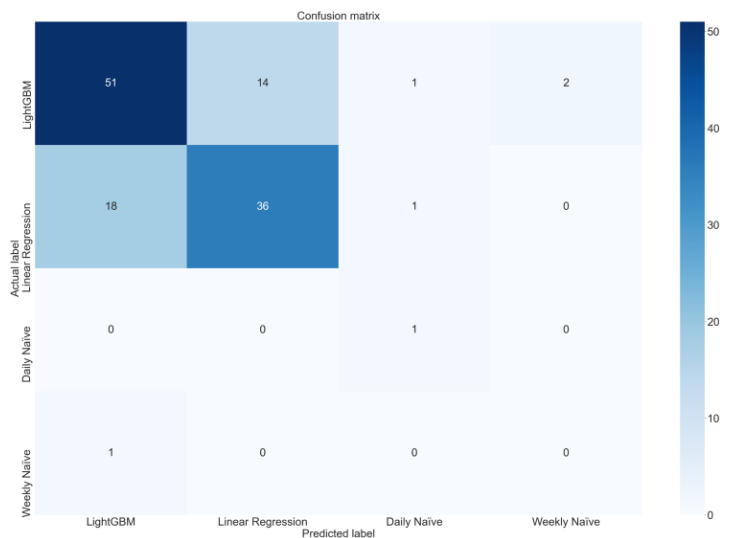

Fig. 8. Confusion matrix of the ifeel classification model.

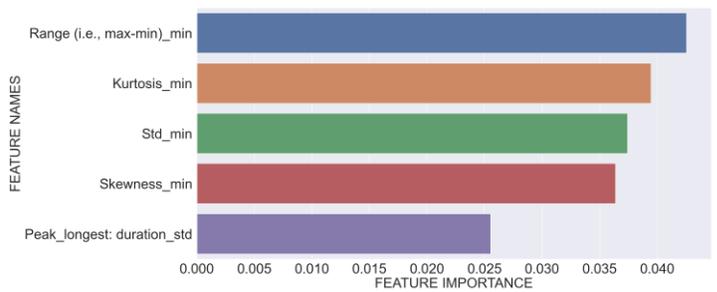

Fig. 11. Top five features of the domain-informed package ifeel.

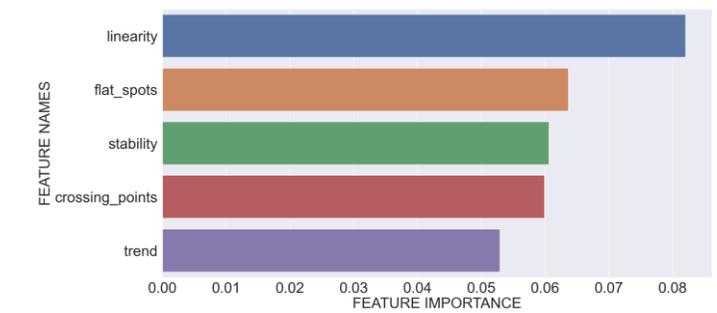

Fig. 12. Top five features of the domain-agnostic package Nixtla.

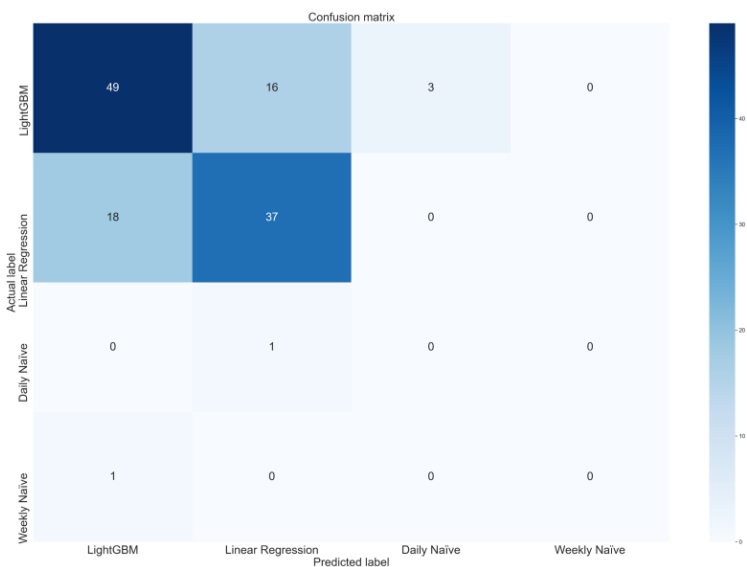

Fig. 9. Confusion matrix of the Nixtla classification model.

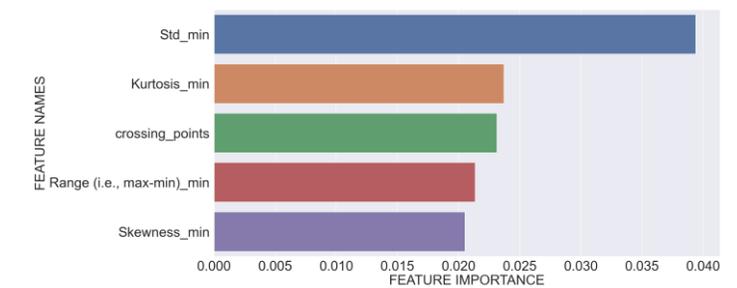

Fig. 13. Top five features of the combination feature matrix.

## VII. Conclusion

Many research studies have been done on forecasting and analysis of building energy consumption at different scales, and over several time horizons, and have reported successful prediction performance. However, predicting the forecastability of the underlying building electricity demand time series is still a challenging gap. Therefore, this research study has proposed a data-driven forecastability framework to address this gap.

The experimental results reported in this research work, show that the proposed forecastability approach is an appropriate framework to make a former estimates about the forecastability of electricity demand time series of such a building, and should therefore be regarded as suitable and optimistic solution for generalization among building science, and data science professionals.

As a future research, we will apply the forecastability framework on buildings that are in different geographic locations. Furthermore, we will explore the possibility of using different data-driven models to predict over and using other models than RF for the forecastability.

## Acknowledgment

This work was supported by Newcastle University and the Engineering and Physics Science Research Council (EPSRC) [grant number EP/S016627/1]: Active Building Centre Project. Hussain Kazmi acknowledges support from Research Foundation – Flanders (FWO), Belgium (research fellowship 1262921N).